\documentclass[journal]{IEEEtran}
\usepackage[T1]{fontenc}
\usepackage[utf8]{inputenc}
\usepackage{authblk}
\usepackage{multirow}
\usepackage{times}
\usepackage{epsfig}
\usepackage{graphicx}
\usepackage{amsmath}
\usepackage{amssymb}
\usepackage{booktabs}
\usepackage{multirow}
\usepackage[labelfont=bf]{caption}                         
\usepackage[subrefformat=parens]{subfig}
\usepackage{enumitem}  
\usepackage{amssymb}
\usepackage[misc]{ifsym}
\usepackage{graphicx}
\usepackage{threeparttable}
\usepackage{ntheorem}
\usepackage{color}
\usepackage{colortbl}
\usepackage{multirow}
\usepackage{amsmath}
\usepackage{comment}
\usepackage{bm}                                            
\usepackage{etoolbox}                                      
\usepackage{url}
\usepackage{nth}
\usepackage{cite}
\usepackage{balance}
\usepackage[bookmarks=false]{hyperref}                     %
\usepackage{courier}                                       
\usepackage{listings}                                      
\usepackage[]{algpseudocode}                               
\algrenewcommand\textproc{\texttt}
\makeatletter\let\c@float@type\relax\makeatother
\makeatletter\let\float@addtolists\relax\makeatother
\usepackage{algorithm}
\usepackage{bbding}
\usepackage{pifont}
\usepackage{wasysym}
\usepackage{makecell}

\hypersetup{
	colorlinks = true,
	citecolor  = blue,
	linkcolor  = blue,
	urlcolor   = blue,
}

\newcommand{\tabincell}[2]{                                
	\begin{tabular}{@{}#1@{}}#2\end{tabular}
}

\makeatletter
\newcommand{\thickahline}{%
	\noalign {\ifnum 0=`}\fi \hrule height 1pt
	\futurelet \reserved@a \@xhline
}
\graphicspath{{./figs/}}
\begin{document}

\title{Non-local Channel Aggregation Network for Single Image Rain Removal}
%

%
\author[1]{Zhipeng Su,~\IEEEmembership{Student Member,~IEEE}}
\author[1]{Yixiong~Zhang,~\IEEEmembership{Member,~IEEE}\thanks{Corresponding author: Yixiong Zhang (email: zyx@xmu.edu.cn)}}
\author[2]{Xiao-Ping Zhang,~\IEEEmembership{Fellow,~IEEE}}
\author[3]{Feng Qi}
\affil[1]{Department of Information Science and Engineering, Xiamen University, China}
\affil[2]{Department of Electrical Computer and Biomedical Engineering Ryerson University, Toronto, ON, Canada}
\affil[3]{Shenyang Institute of Automation, Chinese Academy of Sciences}

\maketitle

\IEEEpeerreviewmaketitle

%

\begin{abstract}
Rain streaks showing in images or videos would severely degrade the performance of computer vision applications. Thus, it is of vital importance to remove rain streaks and facilitate our vision systems. While recent convolutinal neural network based methods have shown promising results in single image rain removal (SIRR), they fail to effectively capture long-range location dependencies or aggregate convolutional channel information simultaneously. However, as SIRR is a highly illposed problem, these spatial and channel information are very important clues to solve SIRR. First, spatial information could help our model to understand the image context by gathering long-range dependency location information hidden in the image. Second, aggregating channels could help our model to concentrate on channels more related to image background instead of rain streaks. In this paper, we propose a non-local channel aggregation network (NCANet) to address the SIRR problem. NCANet models 2D rainy images as sequences of vectors in three directions, namely vertical direction, transverse direction and channel direction. Recurrently aggregating information from all three directions enables our model to capture the long-range dependencies in both channels and spaitials locations. Extensive experiments on both heavy and light rain image data sets demonstrate the effectiveness of the proposed NCANet model.

\end{abstract}


\section{Introduction}

\label{sec:introduction}

\PARstart{B}{ad} weather conditions pose a great threat to our outdoor vision systems, which would lead to severe system performance degradation. 
These systems such as autonomous driving cars, video surveillance systems and flying drones etc, are usually vulnerable, where the developers of the systems usually assume that the feed of data sets to the system are clean images without any contamination.
However, bad weather conditions would cause content deformation like blurring or obstructing the background of the image and even alter the content of the image, and thus serve as a disaster to these vision systems.
Rainy days are one of the most common whether conditions, and when outdoor vision systems meet rain streaks from their inputs, the fluctuations of the image pixel intensities would cause serious confusion to the systems.

\begin{figure}[!t]
	\centering
	\subfloat[Background]{\includegraphics[width=.48\columnwidth]{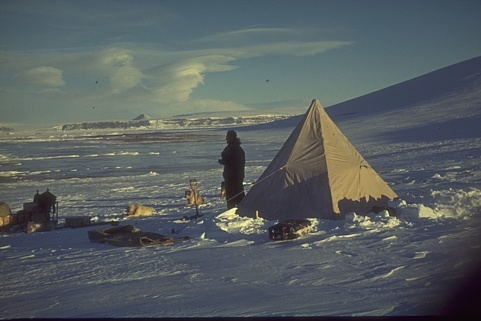}  \label{fig:bg_013}}
	\subfloat[Rainy image]{\includegraphics[width=.48\columnwidth]{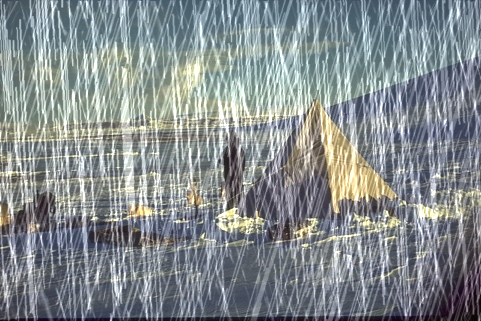} \label{fig:rain_013}} \\
	\subfloat[NCANet]{\includegraphics[width=.48\columnwidth]{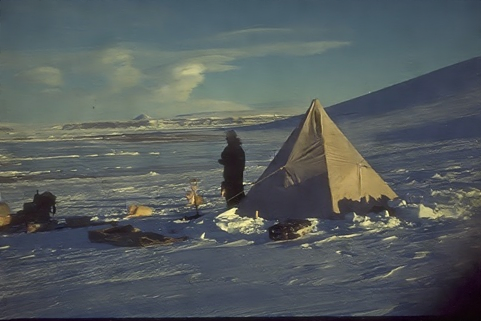} \label{fig:derain_013}}
	\subfloat[RESCAN \cite{cnn_squeeze_2018}]{\includegraphics[width=.48\columnwidth]{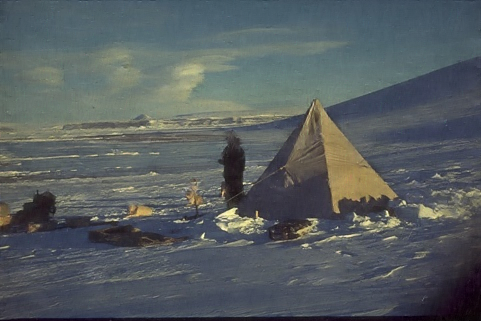} \label{fig:rescan_013}}
	\caption{ 
	    Example deraining results by RESCAN \cite{cnn_squeeze_2018} and NCANet.
	}
	\label{fig:derain_example_1}
\end{figure}

In order to rescue these vision applications like image classification \cite{resnet2016}, image compression \cite{2020tmmcom, 2019tmm}, object detection \cite{maskrcnn2017} and urban scene segmentation \cite{cityscapes2016}, numerous endeavors devote themselves to the development of intelligent rainy/haze image \cite{2020TMM} restoration algorithms.
Image rain removal is known as a ill-posed problem, and there are countless mappings between inputs and outputs, namely between rainy images and rain-free backgrounds.
Therefore, several traditional methods \cite{video_tensor_2017, cnn_density_2018, image_sparse_2015} impose certain constraints or regularizers on rainy images.

\begin{figure*}[th!]
	\centering
	\includegraphics[width=1.6\columnwidth]{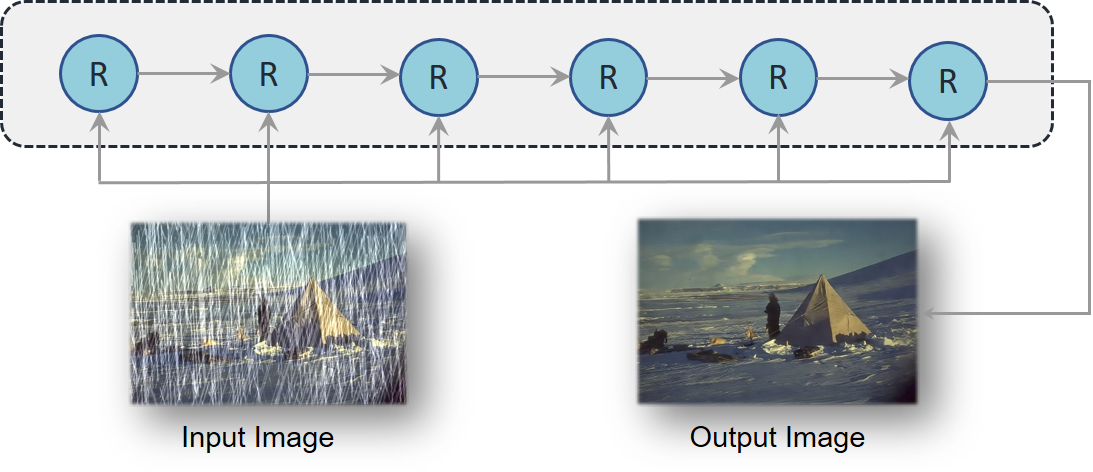}
	\caption{ 
		Overall framework of NCANet: it is built by recurrently connect our building units, R block.
	}
	\label{fig:framework}
\end{figure*}

One main stream of methods focus on recovering rainy images from a sequence of video frames \cite{video_tensor_2017, video_stochastic_2017, video_matrix_2017, video_erase_2018, video_sparse_2018,video_hist_2011,video_phase_2015,video_dynamic_2013,video_lowrank_2015}, as they could leverage the temporal information within these video frames.
These temporal information could provide useful clues and constraints for deraining models.
However, using video sequences for deraining would significantly delay the pre-processing procedure for our computer vision systems, which is usually intolerable for some real-time applications.
The latest image frame fed to the vision system will always delay several frames compared to the latest raw frame, because video based image deraining algorithms need to take a few frames from the video only to process one derained image.
Nevertheless, there is still place for video based deraining algorithm which is generating deraining data sets with high quality in real life scenes \cite{cnn_spatial_2019}.

Another stream of methods pay their attention to single image rain removal (SIRR), which is also the target of our paper.
Compared to video based rain removal, SIRR is more ill-posed and become significantly more challenging, as there is no temporal information available.
In the past decades, researchers proposed numerous patch-based and prior-based methods \cite{image_sparse_2015, image_layer_2016, image_bilayer_2017, image_generalized_2013, image_structural_2014, image_automatic_2011, image_convolutional_2017} to sovle SIRR problem. And Akbari et al \cite{2017Sparse} also consider derain as an image recovery problem such as error concealment \cite{2019Joint}.
Most of these methods impose certain regularizers, or they would try different priors to perform layer separation directly on the images.
However, these prior-based or regularizer-based approaches suffer from over-smoothing the background or under/over deraining the images.
It is easy to understand that these priors or regularizers have limited information to model the rain in natural images.

\begin{figure*}[th!]
	\centering
	\includegraphics[width=1.9\columnwidth]{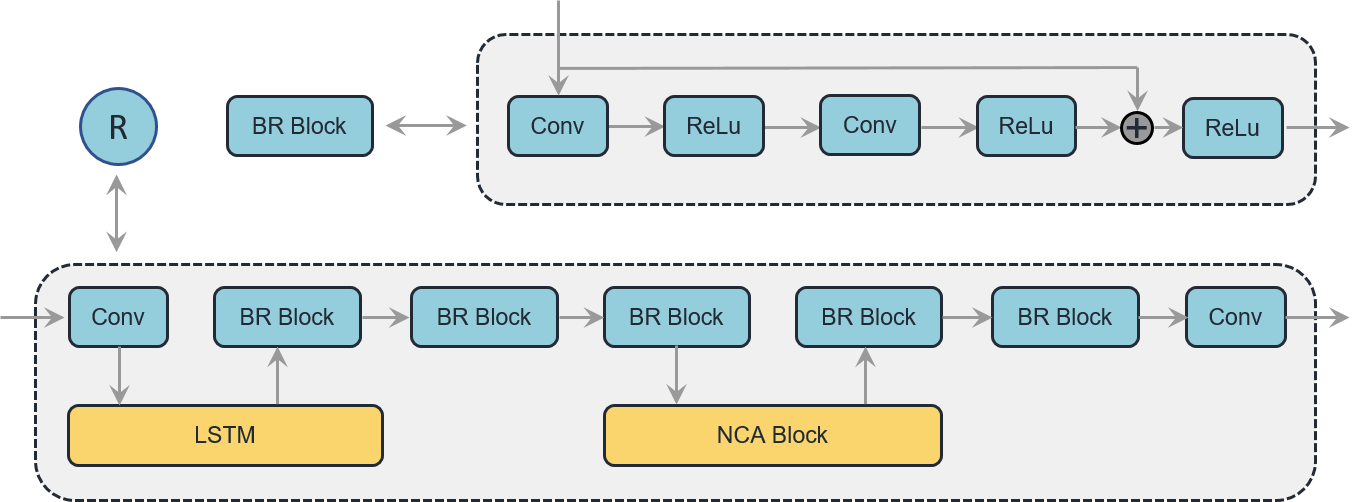}
	\caption{ 
		Details of R block: it contains five BR blocks, two normal convolutional blocks, a LSTM block and a NCA block.
	}
	\label{fig:resblock}
\end{figure*}

Recent years have witnessed rapid growth of deep learning models, especially deep convolutional neural networks (CNN).
With the successful applications of CNN in other low-level vision tasks like super resolution \cite{othercnn_super_2017} and image denoising \cite{othercnn_denoiser_2017}, CNN have also been widely used in SIRR problem \cite{cnn_density_2018, cnn_spatial_2019, cnn_residual_2018, 2020cnntmm,cnn_detail_2017,cnn_generative_2018, cnn_progressive_2019, cnn_squeeze_2018, 2020smoke, cnn_lightweight_2019}.
People usually deploy functional blocks from network structures such as residual nets \cite{resnet2016, cnn_progressive_2019}, dense nets \cite{densenet2017, cnn_residual_2018} and squeeze-and-excitation nets \cite{squeeze2018, cnn_squeeze_2018} to enhance their network.
Among these networks, RESCAN \cite{cnn_squeeze_2018} is the first one to try attention network, where it uses squeeze-and-excitation block to force the network to pay more attention to the important channels.
However, RESCAN fails to incorporate global spatial correlations \cite{nonlocal2018} that could provide long-range dependencies in these rainy images. 
As shown in Fig.~\ref{fig:derain_example_1}, though RESCAN uses channel attention block in their network, their results seem to be more noisy than ours, where our model considers channel and spatial attention simultaneously.
Non-local (NCL) block \cite{nonlocal2018} is a recent poweful attention block that could capture long-range dependencies; however, NCL block will fail when input image is too large as it will consume large memory space.
Also, NCL block ignores channel information which serves as an important clue for image rain removal.

Though traditional methods mentioned above have achieved success of certain degree, there exist several major drawbacks among these methods:

\begin{itemize}
    \item Video based methods could normally get more clean images, but video frames will cause serious delay for real-time outdoor vision systems;
    \item Prior or regularizer-based methods rely on image patches, and they ususally suffer from over-smoothing and under/over derain problem;
    \item Deep CNN-based methods gain better images, but they still ignore important spatial and channel information, and most of them are complicated and hard to replicate.
\end{itemize}
Therefore, it is imperative to build a unified CNN that could solve the aforementioned problems. 
Motivated by these issues, we propose a non-local channel aggregation network (NCANet) that incorporates spatial and channel correlational information simultaneously.
The pipeline of our model is shown in Fig.~\ref{fig:framework}. 
We propose a new rain model that regards the rain layer as the sum of multiple residuals between the original image and the background gained by model, because we believe rain is accumulated to the image,  and there are links between intermediate status.
Based on this new rain model which could better approximate the real rain layer, we develop a novel recurrent neural network for SIRR.
The fundamental building block of our model is called residual block (R block), as is shown is Fig.~\ref{fig:framework} and detailed in Fig.~\ref{fig:resblock}. In each R block, we have one non-local channel aggregation (NCA) block, which enables our model to capture and aggregate the spatial and channel-wise information simultaneously.
To the best of our knowledge, it is the first time that NCA block and recurrent neural network are developed together to perform rain removal task.

Main contributions of our paper are listed as follows:

\begin{itemize}
	\item We propose a new rain model that could better approximate the nature of accumulated rain streaks on the image. Unlike previous rain model that only regards the original image is the sum of background layer and rain layer, we model the rain layer as the sum of multiple residuals between original images and intermediate background images.
	\item We propose a novel unified recurrent convolutional residual network to fit in our rain model, which recurrently run our R block for six times; and in each R block the attention block NCA could simultaneously capture both long-range spatial and channel correlations, which helps our model capture more detailed information for deraining.
	\item Extensive experimental results on both heavy and light rainy images show the superiority of our method compared to other stat-of-the-art methods. Our ablation study shows the stability of our proposed NCA block.
\end{itemize}

\section{Related Works}

\label{sec:related}

There are three types of related works: video based deraining, patch based image deraining and CNN based image deraining.
We will briefly review these three types of deraining works in this section.

\subsection{Video Rain Removal}

For video based rain removal, we could easily discover clues from a sequence of images, as there exists rich temporal information.
Many of these methods achieve reasonably good results, but has nothing to do with the single image deraining task, as there is no more temporal information.
However, video based methods would cause significant delay to modern outdoor vision systems, which is intolerable for applications like autonomous driving.
Still and all, video based methods could help generate clean deraining datasets \cite{cnn_spatial_2019}.
\cite{video_hist_2011} uses a histogram of orientation of rain streaks to capture the characteristics of the rainy images. 
\cite{video_dynamic_2013} performs deraining in a dynamic scene, which is reasonable as in this situation SIRR may fail.
\cite{video_tensor_2017} imposes constraints on rainy images based on certain priors.
\cite{video_lowrank_2015} performs both deraining and desnowing based on the inter-frame temporal information and low-rank regularization.
\cite{video_sparse_2018} also applies convolutional sparse coding to approximate background rain-free images.

\subsection{Optimization based Single Image Rain Removal}

Traditional methods here denote those do not use deep neural network as their backbone methods.
Single-image rain removal (SIRR) is much more challenging than video based image deraining, as SIRR is highly ill-posed and lacks of enough clues for our models to recover the rainy images.
Thus, researchers resort to prior and regularizer based methods \cite{image_sparse_2015, image_layer_2016, image_bilayer_2017, image_generalized_2013, image_structural_2014, image_automatic_2011, image_convolutional_2017}.
These methods usually decompose the original image into different patches overlapping with each other, assuming that original images could be composed by sum of the background image layer and rainy image layer.
Also, they deploy certain numerical optimization methods, such as ADMM \cite{admm_2011} and L-BFGS \cite{lbfgs_1997} to solve the proposed optimization problem.
\cite{image_automatic_2011} discovers high-frequency information from rainy layers and applies sparse coding methods to separate rain layers from background.
\cite{image_sparse_2015} developed a different rain model than \cite{image_automatic_2011}, and also applies sparse coding to derain the image.
\cite{image_layer_2016} imposes different regularization on background layer and rain layer respectively.
\cite{image_bilayer_2017} improves the model in \cite{image_layer_2016} by imposing three different priors, which are sparse prior, rain direction prior and rain layer prior.
\cite{image_structural_2014} further explores structural similarity for efficient deraining.
All these methods assumes a prior information to their developed optimization; however, these priors or regularizers only contain limited information for image deraining.
Thus, these prior based methods are prone to over-smooth or under/over derain.

\subsection{Deep Learning based Single Image Rain Removal}

Deep learning shows a strong boost for both high-level computer vision application \cite{face_2015, faster_2015} and  low-level computer vision tasks \cite{othercnn_super_2017, othercnn_denoiser_2017, cpp_2020}, and so do our deraining tasks \cite{cnn_density_2018, cnn_spatial_2019, cnn_residual_2018, cnn_joint_2017,cnn_detail_2017,cnn_generative_2018, cnn_progressive_2019, cnn_squeeze_2018, cnn_Clearing_2017, cnn_lightweight_2019}.
These deep learning models usually learn a mapping between input rainy images and output rain-free images, and achieve superiority performance.
Residual blocks \cite{resnet2016}, dense blocks \cite{densenet2017}, squeeze-and-excitation \cite{squeeze2018} are commonly used building units in these deep learning models.
\cite{cnn_squeeze_2018} uses squeeze-and-excitation block to capture the importance between different channels, but ignores the spatial correlations in the rainy images.
\cite{cnn_progressive_2019} applies residual blocks with iterative training procedures.
\cite{cnn_density_2018} deploy multi-branch deep neural network structure, which aims at capturing rain density in one branch and combining all information on another branch.
\cite{cnn_detail_2017} uses detailed image layer as input and gained a non-trivial performance boost.
Also, adversarial network \cite{cnn_generative_2018} is also used to compensate rainy part of the input image.
In comparison to all these CNN models, we simply wrap our model into a single residual block with non-local channel aggregation block incorporated, and perform concatenation and training similar to recurrent neural network \cite{rnn_2015}.

\section{Rain Model}

\label{sec:network}

In this section, we will introduce our new rain model.
In the past works, people usually applied a linear summation model of deraining, where the model assumes the original image (a.k.a the input image) is the sum of rain-free background image layer and rainy image layer.
This traditional blending model is as follows:

\begin{equation}
	\mathbf{O} = \mathbf{B} + \mathbf{R},
	\label{eqn:traditional_blend}
\end{equation}
where $\mathbf{O}$ is the original observed image, $\mathbf{B}$ is the rain-free background image, and $\mathbf{R}$ is the layer with rain streaks.
Usually, traditional methods remove rain layers from original observed image, that is remove $\mathbf{R}$ from $\mathbf{O}$, to get rain-free image.

We argue that building a simple deraining model merely based on Eqn.~\eqref{eqn:traditional_blend} is far from solving this problem.
On the one hand, as stated in \cite{cnn_density_2018}, it is a important clue for our model to understand how much rain is there in the image, or the deraining model is prone to suffer from under/over deraining problem due to the incorrect estimation of rain density.
On the other hand, the directions, shapes and distributions of rain vary a lot under different raining conditions, it is a good idea but not the best to model rain in Eqn.~\eqref{eqn:traditional_blend}.

To capture these complex rain variations and further reduce the burden of deraining model, our new rain model regards the rain layer as a linear combination of several residuals between these intermediate images $\mathbf{B}_i$:
\begin{equation}
	\sum_{i=3}^{N} (\mathbf{B}_{i} - \mathbf{B}_{i-1}) + \mathbf{B}_{2} = \mathbf{R} + \mathbf{O},
	\label{eqn:new_blend1}
\end{equation}
where $N$ is the number of intermediate rain-removed background images (consider $\mathbf{O}$ as one of the intermediate image), and $\mathbf{B}_i$ is the $i_{th}$ intermediate background image.
By setting  $\mathbf{O} =  \mathbf{B}_1$, we could further simplify the rain model as follows:
\begin{equation}
	\sum_{i=2}^{N} (\mathbf{B}_{i} - \mathbf{B}_{i-1}) = \mathbf{R}.
	\label{eqn:new_blend2}
\end{equation}

In order to fit our new rain model, we developed an iterative deraining model NCANet, which will be discussed in the next few sections.

\begin{figure*}[th!]
	\centering
	\includegraphics[width=1.9\columnwidth]{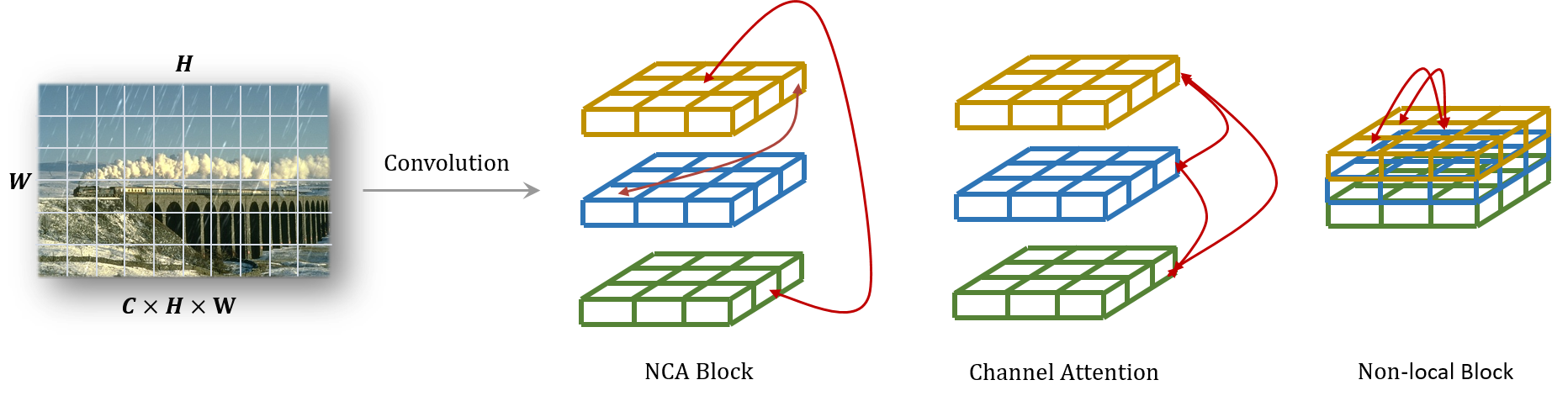}
	\caption{ 
		Attention block comparison: from left to right are our NCA block, channel attention block \cite{cnn_squeeze_2018}, and non-local block \cite{nonlocal2018} (a.k.a spatial attention block). 
	}
	\label{fig:nca1}
\end{figure*}

\section{Derain Models}

In this section, we will introduce detailed building units of our non-local channel aggregation network (NCANet). 
The overall framework of NCANet can be found via Fig.~\ref{fig:framework}, and details of the residual block (R block) is shown in Fig.~\ref{fig:resblock}. We will introduce these building units from bottom to up, where the basic units such as non-local channel aggregation block (NCA Block), basic residual blocks (BR block), LSTM \cite{lstm1997} block will be introduced one by one, followed by the overall framework.

\subsection{Non-local Channel Aggregation Block}
\begin{figure}[th!]
	\centering
	\includegraphics[width=1.\columnwidth]{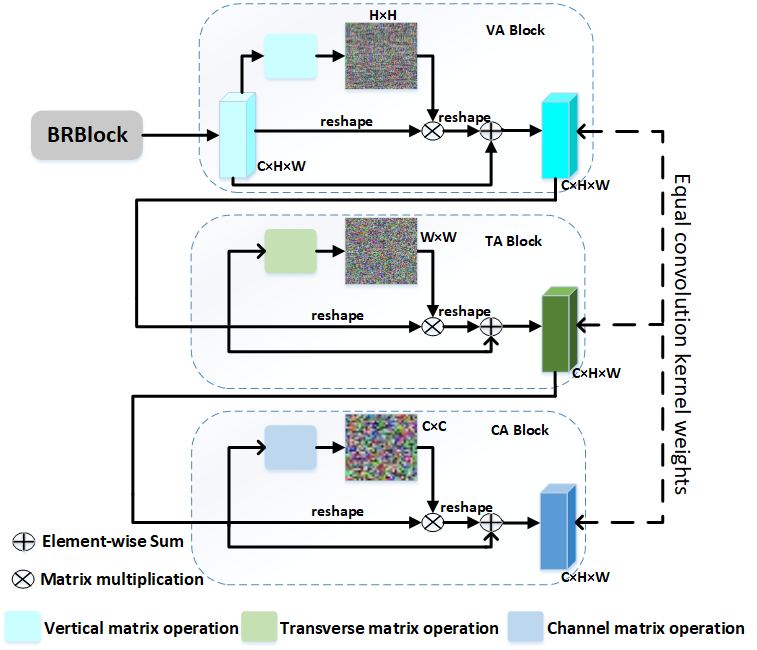}
	\caption{ 
		Visualization of the NCA block, which exhibit attention map of three different dimensions. The similarity between one pixel and all other pixels matrix weight (attention map), which have higher degree of resemblance learned, the greater softmax value acquired after the matrix operation.
	}
	\label{fig:nca4}
\end{figure}

As shown in Fig.~\ref{fig:nca1}, the image can be decomposed in three layers in the feature space $C\times H\times W$, where $C$ is the number of channels in the feature map after convolutions, and $H$,$W$ are the height and width of the image respectively. 
Basically, traditional methods \cite{nonlocal2018, squeeze2018} could only capture part of the correlation in these layers, resulting in the loss of important information.
\cite{nonlocal2018} propose a so-called non-local (NCL) bock, as shown in the non-local block in Fig.~\ref{fig:nca1}, and the major drawback of this method is that it doesn't capture the channel attention, and the attention produce by the NCL block is sometime huge and cannot be fed into the memory, which is the case in our deraining application. 
\cite{squeeze2018, cnn_squeeze_2018} utilizes the channel attention, the principle of which is shown in the channel attention part in Fig.~\ref{fig:nca1}.
Though channel attention captures the long-range dependency of these channels, however, it ignores important clues for spatial relationships. As shown in Fig.~\ref{fig:nca4}, In order to combine spatial and channel information, we address these issues by iteratively processing the three dimensions in turn.

As mentioned in introduction, CNN based methods suffer from capturing important rain streak clues hidden in the rainy images.
These clues include spatial location correlations between different rain streaks. Channel correlations inside CNN models which could enable better rain layer separation ability. It can also distinguish different rain densities in the same background scene.
\cite{cnn_squeeze_2018} tries to separate the rain using channel attention block called squeeze-and-excitation block, however, this block does not consider the spatial correlations and thus result in the loss of important rainy layer clues.
Other methods do not even consider these correlations, though \cite{cnn_density_2018} does consider the rain density but require additional rain density labels for model training, which is not applicable in most application cases.
Thus, we propose our non-local channel aggregation block (NCA Block) as shown in Fig.~\ref{fig:nca4} to resolve these issues.
Our NCA block could capture long-range dependencies between rain streaks in different locations and also channel correlations when training our CNN models, which provide excellent performance in separating rain layers.
We will see more detailed results in the experimental part of the paper.

It is easy to generate an attention map for every one-by-one correlation; however, take the NCA block in the Fig.~\ref{fig:nca1} as an example.
If we set $H = 512$, $W = 1024$ and $C = 256$, and each cell in the attention map consumes 1 byte, then the attention map size will be around $1.8 \times 10^{16} $ bytes, which will consume unreasonable amount of resources to compute.
Luckily, our NCA block adopts recurrent vector attention techniques that could recurrently perform attention on three directions, i.e. vertical direction, transverse direction and channel direction, and finally approximate simultaneously spatial and channel attention together.

\begin{figure}[th!]
	\centering
	\includegraphics[width=.98\columnwidth]{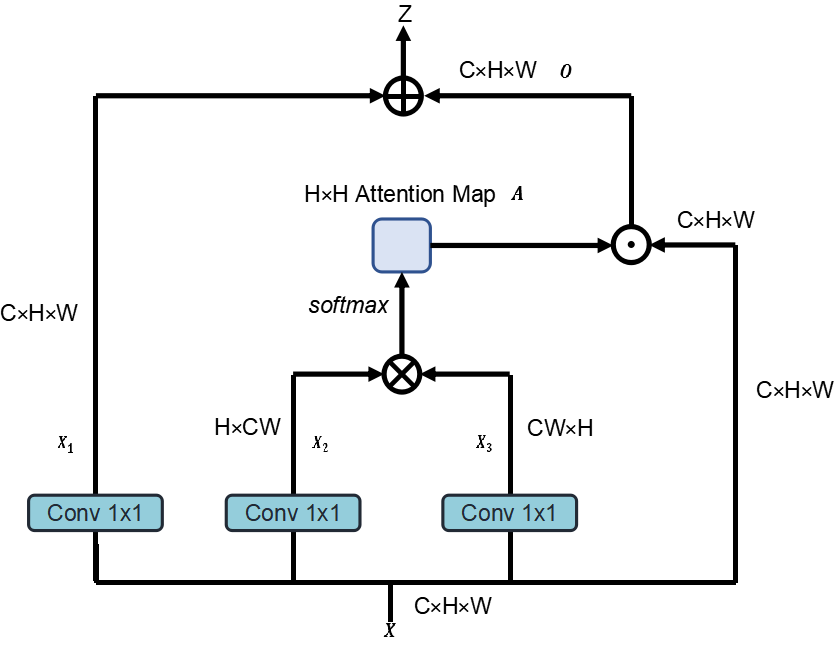}
	\caption{ 
		Sub block example of NCA block, namely VA block, which reshape input tensor layer permutation to $H\times CW$ for the $H\times H$ attention maps by the  matrix multiplication.
	}
	\label{fig:ca}
\end{figure}

\begin{figure}[th!]
	\centering
	\includegraphics[width=.98\columnwidth]{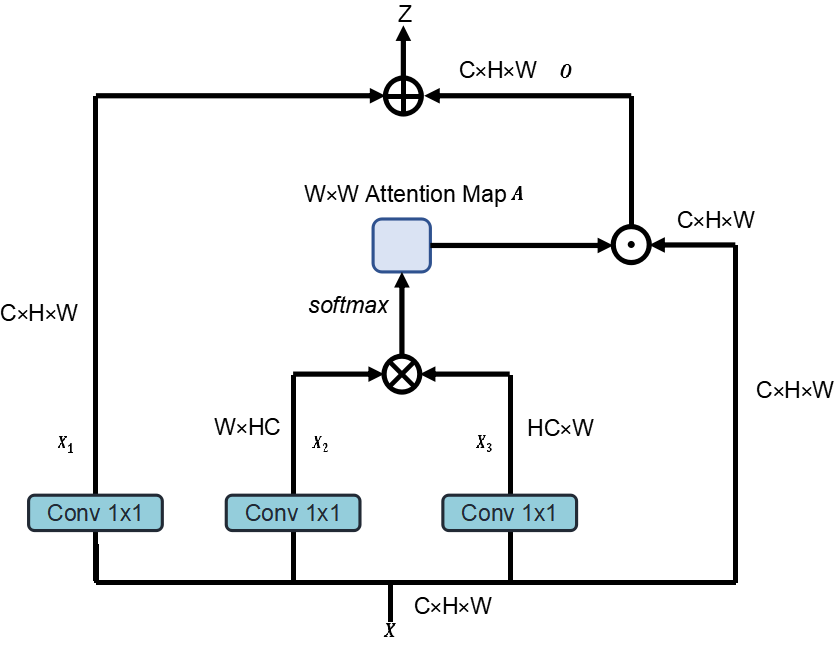}
	\caption{ 
	Sub block example of NCA block, namely TA block, which reshape VA output tensor layer permutation to $W\times HC$ for the $W\times W$ attention maps by the  matrix multiplication.
	}
	\label{fig:va}
\end{figure}

\begin{figure}[th!]
	\centering
	\includegraphics[width=.98\columnwidth]{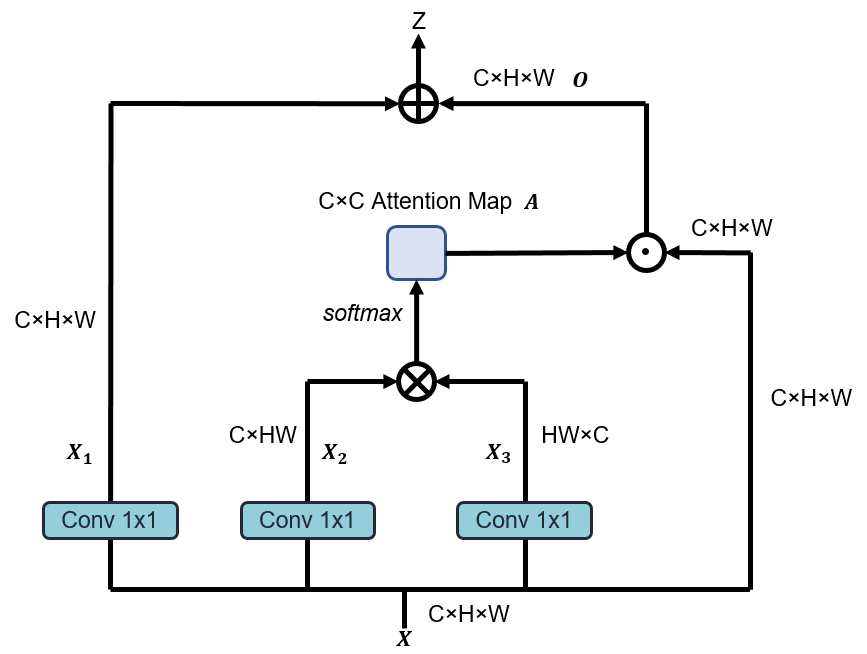}
	\caption{ 
	Sub block example of NCA block, namely CA block, which reshape TA output tensor layer permutation to $C\times HW$ for the $C\times C$ attention maps by the  matrix multiplication.
	}
	\label{fig:ta}
\end{figure}

\begin{figure}[th!]
	\centering
	\includegraphics[width=.98\columnwidth]{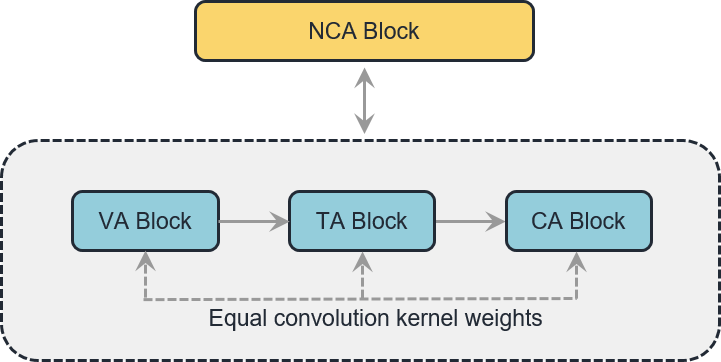}
	\caption{ 
		NCA block consist of three sub blocks, which denotes the three directions in intermediate tensor, i.e. vertical direction, transverse direction and channel direction, and the correlation of these directions are captured by VA block, TA block and CA block respectively.
	}
	\label{fig:nca2}
\end{figure}

NCA block is shown in Fig.~\ref{fig:nca2}, and is composed of three sub blocks, which are vertical attention block (VA blcok), transverse attention block (TA block), and channel attention (CA block) block respectively.
NCA block could capture long-range spatial correlations by recurrently perform vertical attention and transverse attention, and with final channel attention, NCA block could capture one-by-one correlations of each element pair in the $C\times H \times W$ tensor.

For each sub block, we take CA block as an example. 
We have input tensor $X \in \mathbb{R^{C\times H \times W}}$, where $C$ denotes the number of channels of the tensor, $H$ denotes the height of the tensor, and $W$ denotes the width of the tensor.
After $X$ go through three one-dimensional convolution operators separately, we could get three new tensors, which are denoted by $X_1, X_2, \text{and} X_3$ (they have the same size as $X$).
Then, $X_2$ and $X_3$ are reshaped to $X_2 \in \mathbb{R}^{C\times HW}$ and $X_3\in \mathbb{R}^{HW\times C}$, where $HW$ denotes their product.
Thus the whole procedure of CA block could be summarized as the following equation:
\begin{equation}
	Z =  X_1 + \alpha \text{SM}(X_2X_3)X,
	\label{eqn:ca}
\end{equation}
where $\alpha$ is the weight parameter of the attended tensor, which will be updated through back-propagation automatically, and $\text{SM}()$ is the softmax function that would update the matrix row by row.
Other VA and CA blocks follow the same principle as CA block, but they vary in the tensor permutation and attention maps.

The reason one pixel of the output of NCA block could gather information from all other pixels is because: 
\begin{itemize}
	\item A cell in tensor will get a linear combination of all other cells in its column after the VA block
	\item The cell will further get another linear combination of all other cells in its row after the TA block; please note that all other cells in this row have already gathered information from all other cells in their own columns.
	\item The same procedure go for CA block; so after all three blocks, the information of each cell will be propagated to all other cells, thus our NCA block could simultaneously capture the long-range dependencies in spatial and channel correlations.
\end{itemize}

\subsection{Residual Block and Overall NCANet Framework}

Besides the powerful NCA block, we also deploy basic residual blocks (BR block) and LSTM blocks to enhance the whole NCANet.
ResNet \cite{resnet2016} vastly increase the depth of deep neural networks, and has become a basic foundation block for recent computer vision applications.
Also, the efficiency and effectiveness of residual blocks have also been verified by other deraining models in the literature \cite{cnn_detail_2017, cnn_joint_2017, cnn_spatial_2019}.
In contrast to previous methods, we insert additional NCA block between these residual blocks, where NCA block could not only digest the feature maps from previous residual blocks but also provide spatial and channel attended feature map for the subsequent residual blocks.

The overall structure of our residual block is simple, as shown in the bottom panel of Fig.~\ref{fig:resblock}, in which we have five BR blocks, two plain convolution block in the front and back of the residual block, one plain LSTM \cite{lstm1997} block between the first convolution block and the first BR block, and one NCA block between the third BR block and the fourth BR block.
The BR block, as shown in the top panel of Fig.~\ref{fig:resblock}, comprise two consecutive Conv-Relus, a sum operator, followed by ReLu operator \cite{relu2010}.
 
For the overall NCANet, we adopt the rain model in Eqn.~\eqref{eqn:new_blend2}, with iterated processing of our rain-free background image, as shown in Fig.~\ref{fig:framework}.
We have one residual block, and the residual block will be processed recurrently.
Each round, our residual block will get both original rainy image and image processed by last block as input.
After six iteration, we will get our final result.
The superiority of rain model in Eqn.~\eqref{eqn:new_blend2} and the proposed deraining model lies in that with such modeling, our model could be more adaptive to rain under different conditions. As no matter in what condition, our model could process the rainy image iteratively and recurrently.

\section{Experimental Results}

\label{sec:experiment}

In this section, we first conduct ablation study to verify the impact on different locations of putting the NCA block. 
Second evaluate our NCANet quantitatively and qualitatively and conduct comparison study on other state-of-the-art deraining methods. Finally use Pytorch \cite{pytorch2017} to implement our NCANet, and a machine with two Tesla P40 GPUs to train our model.

Structural Similarity Index (SSIM) loss \cite{ssim2004} is verified in several literature \cite{cnn_residual_2018, cnn_progressive_2019} is in deraining or denoising tasks, and thus in our NCANet we also choose SSIM loss to construct our objective function. 
For every dataset, the training input is constructed with a batch size of 8, and each frame of the batch is a $100 \times 100$ image patch.
We use ADAM \cite{adam2014} to train our NCANet, where the initial learning rate is set as $1\times 10^{-3}$ and ends after 100 epochs.

\begin{figure*}[!th]
	\centering
	\subfloat[Rainy Image]{\includegraphics[width=.45\columnwidth]{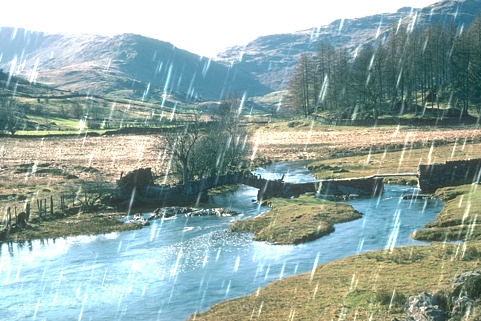}  }
	\subfloat[Background]{\includegraphics[width=.45\columnwidth]{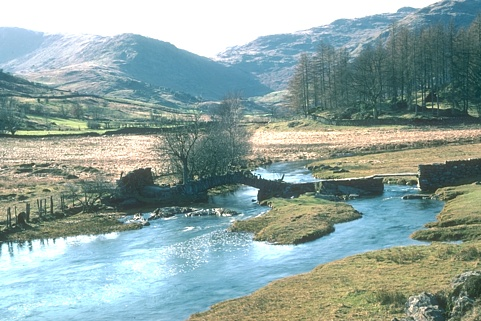}  }
	\subfloat[$\text{NCANet}_1$]{\includegraphics[width=.45\columnwidth]{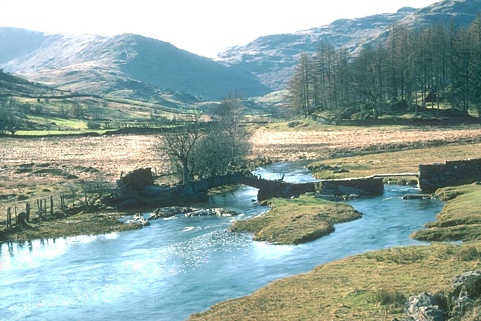} } 
	\subfloat[$\text{NCANet}_2$]{\includegraphics[width=.45\columnwidth]{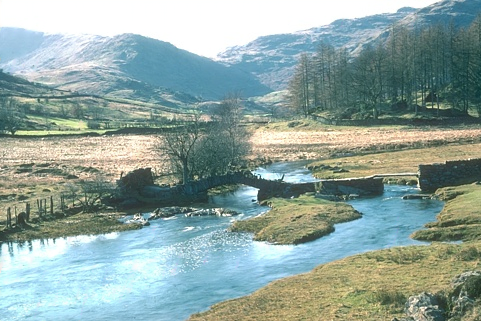} } \\
	\subfloat[$\text{NCANet}_3$]{\includegraphics[width=.45\columnwidth]{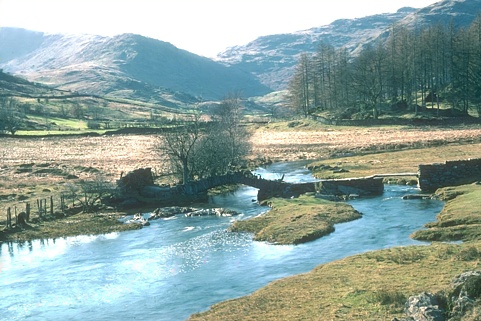} }
	\subfloat[$\text{NCANet}_4$]{\includegraphics[width=.45\columnwidth]{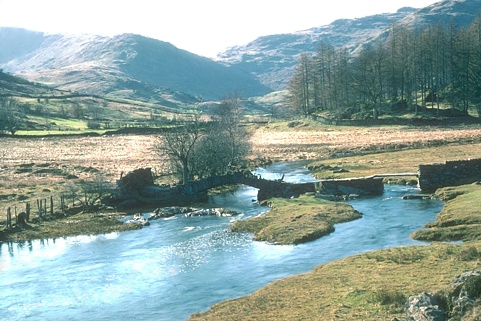} }
	\subfloat[$\text{NCANet}_5$]{\includegraphics[width=.45\columnwidth]{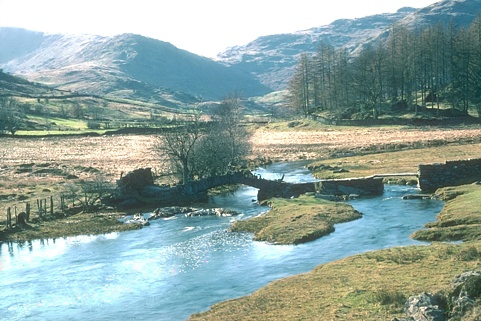} }
	\caption{ 
		Qualitative comparison among NCANets with different NCA block positions on an image from Rain100L.
	}
	\label{fig:ablation_study}
\end{figure*}

%
\begin{figure*}[!th]
	\centering
	\subfloat[Rainy Image\newline{37.2396/0.9445}]{\includegraphics[width=.38\columnwidth]{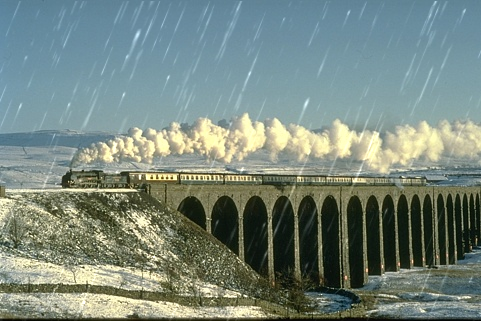}  }
	\subfloat[ Background\newline{Inf/1}]{\includegraphics[width=.38\columnwidth]{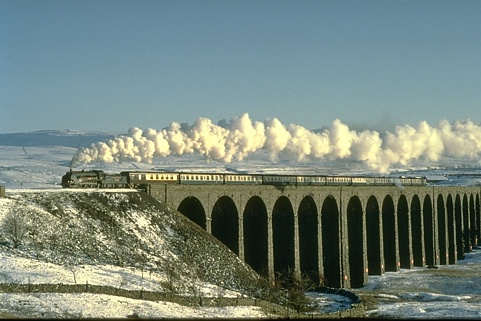}  }
	\subfloat[ JORDER\cite{cnn_joint_2017}\newline{42.8710/0.9929}]{\includegraphics[width=.38\columnwidth]{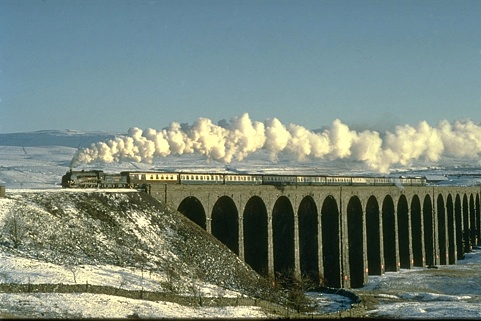} }
	\subfloat[ PReNet\cite{cnn_progressive_2019}\newline{43.4570/0.9936}]{\includegraphics[width=.38\columnwidth]{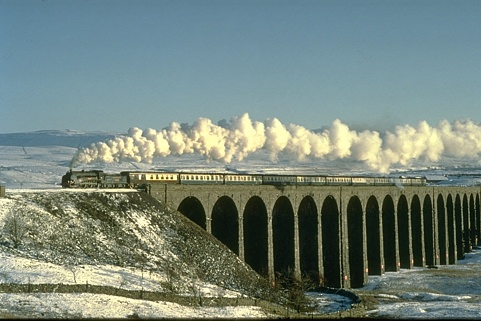} } 
	\subfloat[Our NCANet\newline{\textbf{44.1213/0.9946}}]{\includegraphics[width=.38\columnwidth]{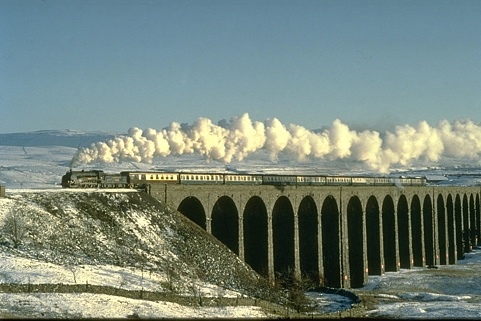} } \\
	\subfloat[Rainy Image\newline{27.6344/0.7853}]{\includegraphics[width=.38\columnwidth]{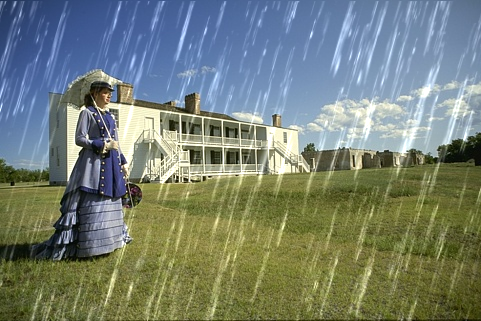}  }
	\subfloat[ Background\newline{Inf/1}]{\includegraphics[width=.38\columnwidth]{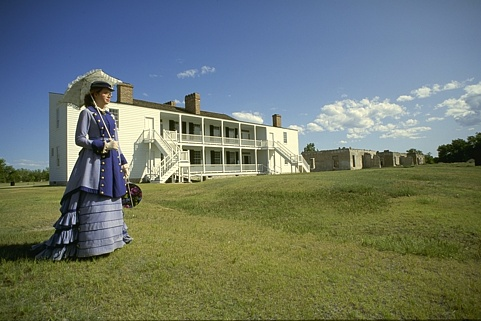}  }
	\subfloat[ JORDER\cite{cnn_joint_2017}\newline{37.5670/0.9664}]{\includegraphics[width=.38\columnwidth]{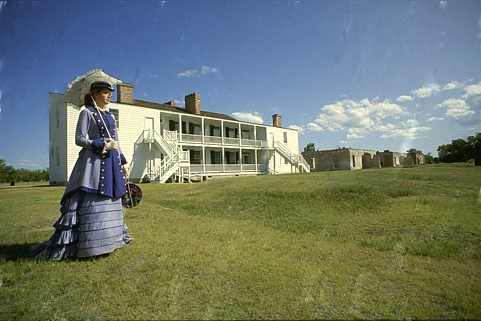} } 
	\subfloat[ PReNet\cite{cnn_progressive_2019}\newline{38.7739/0.9737}]{\includegraphics[width=.38\columnwidth]{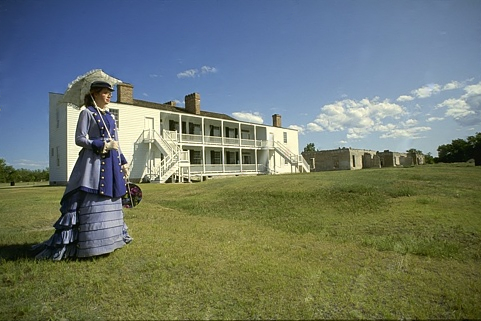} } 
	\subfloat[Our NCANet\newline{\textbf{39.38/0.9766}}]{\includegraphics[width=.38\columnwidth]{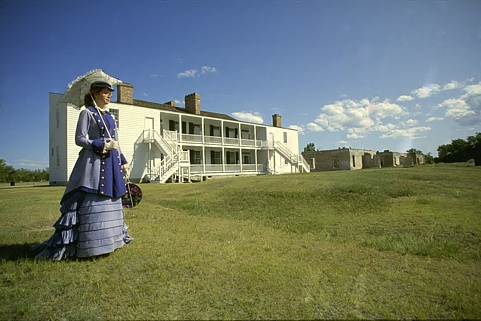} }
	
	\caption{ 
		Qualitative comparison among NCANet and other state-of-the-art methods on images from Rain100L.
	}
	\label{fig:rain100l}
\end{figure*}

\begin{figure*}[!th]
	\centering
	\subfloat[Rainy Image]{\includegraphics[width=.38\columnwidth]{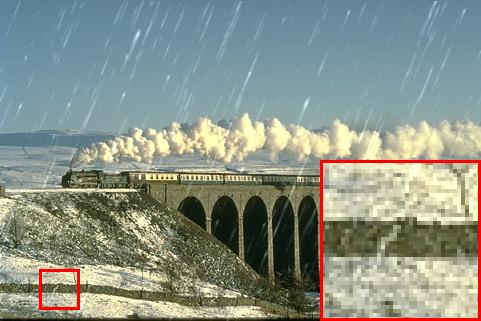}  }
	\subfloat[Background\newline{Inf/1}]{\includegraphics[width=.38\columnwidth]{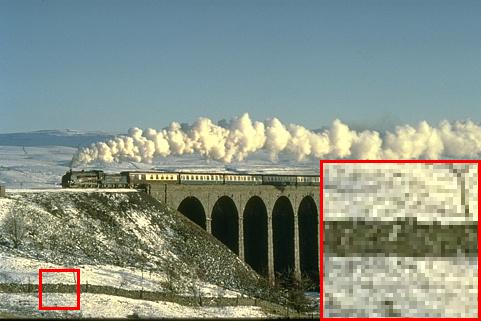}  }
	\subfloat[JORDER\cite{cnn_joint_2017}]{\includegraphics[width=.38\columnwidth]{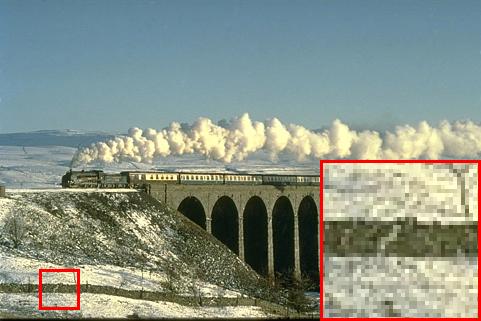} }
	\subfloat[PReNet\cite{cnn_progressive_2019}]{\includegraphics[width=.38\columnwidth]{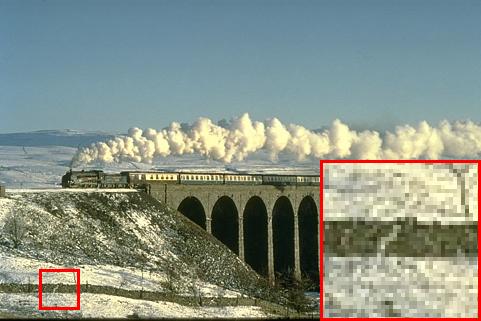} } 
	\subfloat[Our NCANet]{\includegraphics[width=.38\columnwidth]{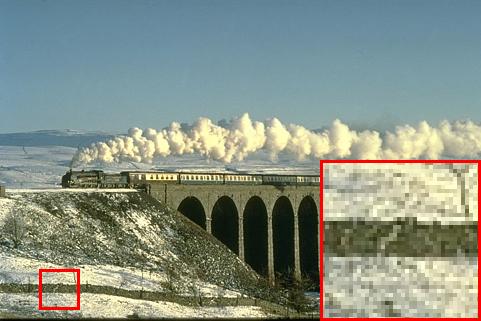} } \\
	\subfloat[Rainy Image]{\includegraphics[width=.38\columnwidth]{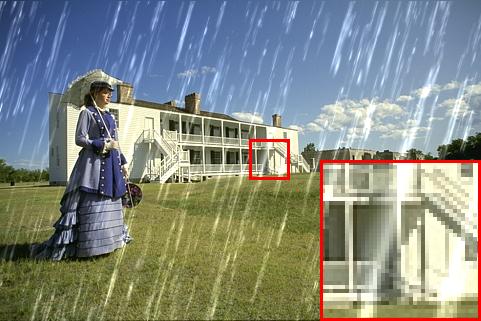}  }
	\subfloat[Background\newline{Inf/1}]{\includegraphics[width=.38\columnwidth]{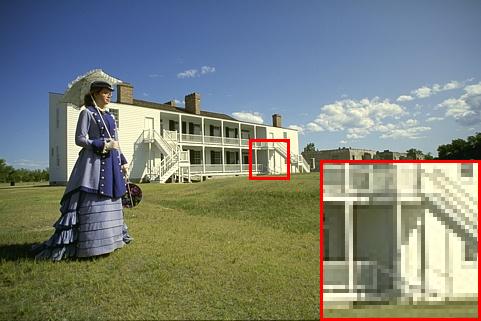}  }
	\subfloat[JORDER\cite{cnn_joint_2017}]{\includegraphics[width=.38\columnwidth]{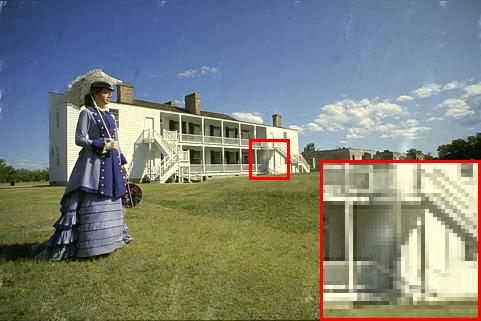} } 
	\subfloat[PReNet\cite{cnn_progressive_2019}]{\includegraphics[width=.38\columnwidth]{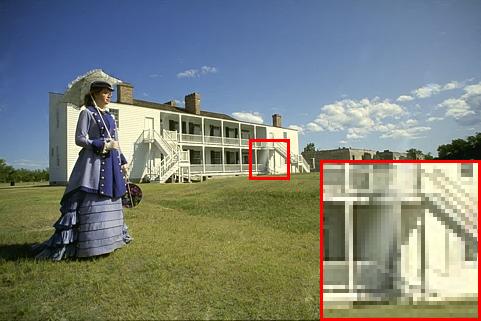} } 
	\subfloat[Our NCANet]{\includegraphics[width=.38\columnwidth]{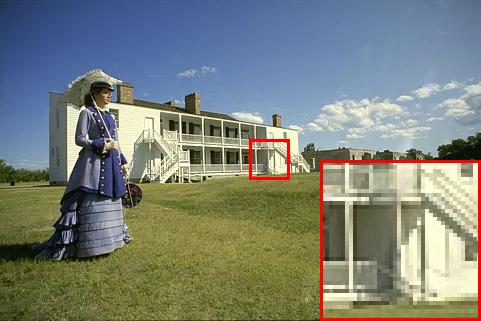} }
	
	\caption{ 
		Enlarged local details for qualitative comparison from Rain100L.
	}
	\label{fig:rain11}
\end{figure*}

\subsection{Ablation Studies}

We conduct of our ablation study on a dataset with 200 training images and 100 testing image called Rain100L \cite{cnn_joint_2017}.
The only parameter in our NCA block is the $\alpha$ parameter; however, $\alpha$ could be learned through the training procedure by back-propagation.
Nevertheless, we are curious about the impact of the position of NCA block in the residual block, and thus we conduct six experiments on different positions of NCA block.

Specifically, we have five positions for our NCA block, which are positions after each BR block respectively. 
In the experiment we call these NCANet with different NCA block positions as $\text{NCANet}_1$, $\text{NCANet}_2$, $\text{NCANet}_3$, $\text{NCANet}_4$, and $\text{NCANet}_5$.
It can be seen from Table~\ref{table:ablation_study}, the quantitative quality of deraining stays the same.
And that's the contribution of our rain model in Eqn.~\ref{eqn:new_blend2}, when we iteratively perform our rain removal procedure, where the NCA block will also involve in these iterations.
Therefore, despite different positions, NCA block will still capture the long-range spatial and channel dependencies simultaneously.

\begin{table}[!hb]
	\centering
	\caption{Without using NCA and LSTM block, and quantitative comparison of NCANet variants: $\text{NCANet}_1$, $\text{NCANet}_2$, $\text{NCANet}_3$, $\text{NCANet}_4$, and $\text{NCANet}_5$ denotes the NCANet with NCA block after the 1st, 2nd, 3rd, 4th and 5th BR block.}
	\vspace{2ex}
	\resizebox{0.5\columnwidth}{!}
	{
		\begin{tabular}{c|c|c}
			\hline \hline
			\tabincell{c}{Model}     &
			\tabincell{c}{PSNR}      &
			\tabincell{c}{SSIM}      \\
			\hline
			$\text{-NCA-LSTM}$     & 36.8396       & 0.976         \\
			$\text{-NCA}$     		& 37.4764       & 0.979         \\
			$\text{-LSTM}$     		& 38.0173       & 0.980         \\	
			$\text{NCANet}_1$        & 38.5011       & 0.982         \\
			$\text{NCANet}_2$        & 38.4775       & 0.981         \\
			$\text{NCANet}_3$        & \textbf{38.5059}       & \textbf{0.982}         \\
			$\text{NCANet}_4$        & 38.4121       & 0.981         \\
			$\text{NCANet}_5$		 & 38.3912       & 0.980         \\
			\hline \hline
		\end{tabular} 
	}
	\label{table:ablation_study}	
\end{table}

\begin{figure*}[!h]
	\centering
	\subfloat[Rainy Image]{\includegraphics[width=.5\columnwidth]{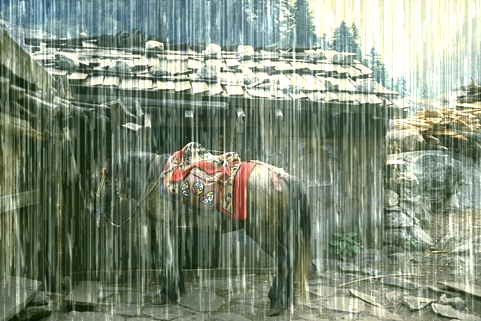}  }
	\subfloat[Background]{\includegraphics[width=.5\columnwidth]{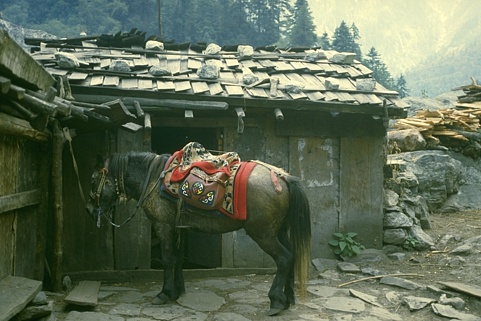}  }
	\subfloat[JORDER\cite{cnn_joint_2017}]{\includegraphics[width=.5\columnwidth]{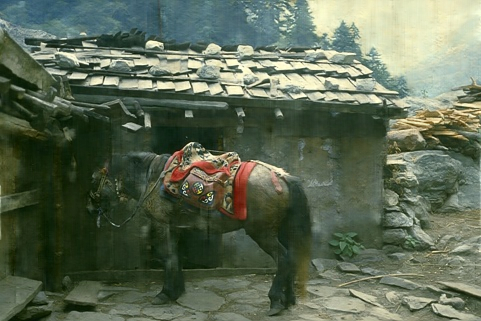} }  \\
	\subfloat[RESCAN\cite{cnn_squeeze_2018}]{\includegraphics[width=.5\columnwidth]{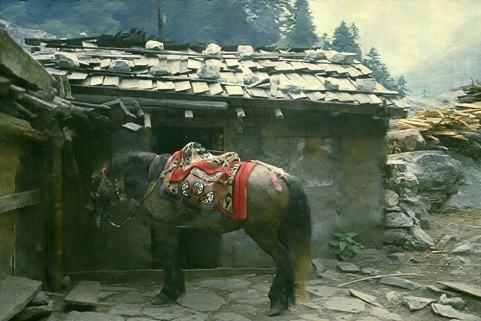} } 
	\subfloat[PReNet\cite{cnn_progressive_2019}]{\includegraphics[width=.5\columnwidth]{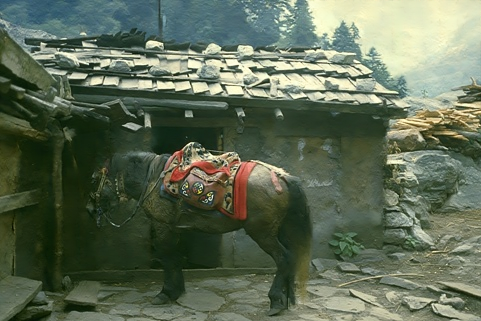} } 
	\subfloat[Our NCANet]{\includegraphics[width=.5\columnwidth]{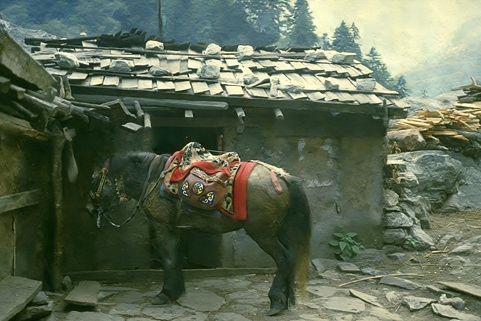} }  \\
	\subfloat[Rainy Image]{\includegraphics[width=.5\columnwidth]{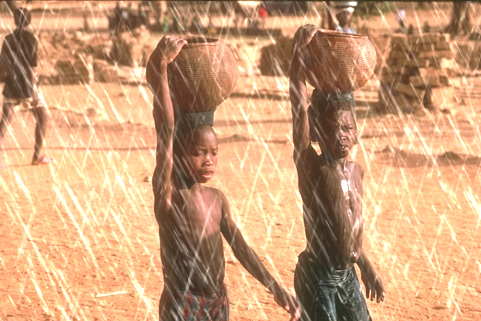}  }
	\subfloat[Background]{\includegraphics[width=.5\columnwidth]{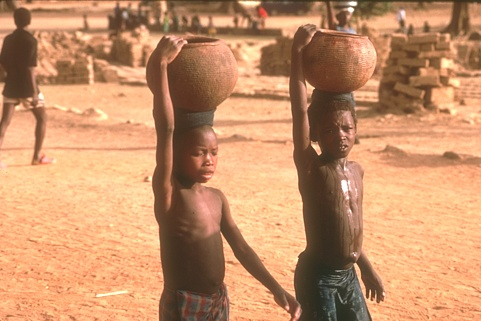}  }
	\subfloat[JORDER\cite{cnn_joint_2017}]{\includegraphics[width=.5\columnwidth]{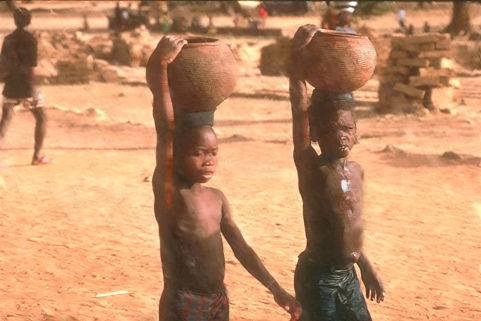} }  \\
	\subfloat[RESCAN\cite{cnn_squeeze_2018}]{\includegraphics[width=.5\columnwidth]{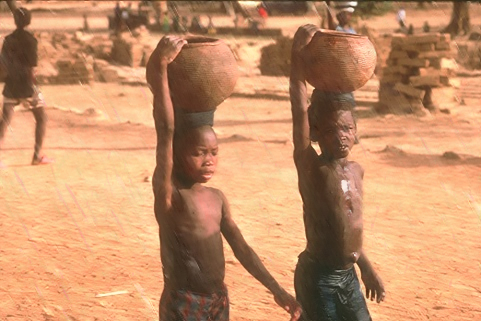} } 
	\subfloat[PReNet\cite{cnn_progressive_2019}]{\includegraphics[width=.5\columnwidth]{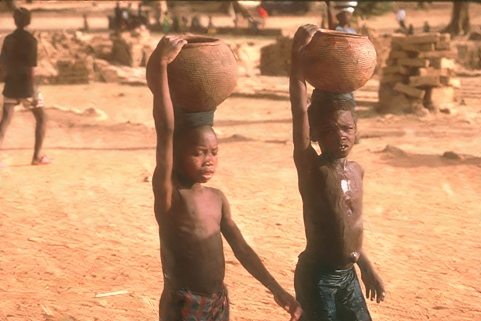} } 
	\subfloat[Our NCANet]{\includegraphics[width=.5\columnwidth]{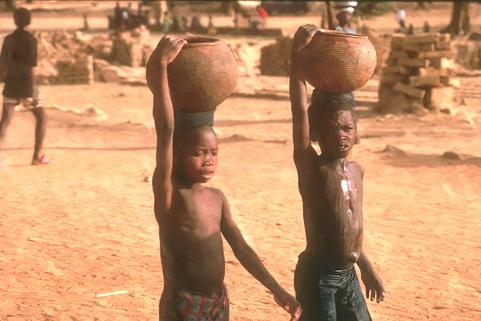} }  \\ 
	\subfloat[Rainy Image]{\includegraphics[width=.5\columnwidth]{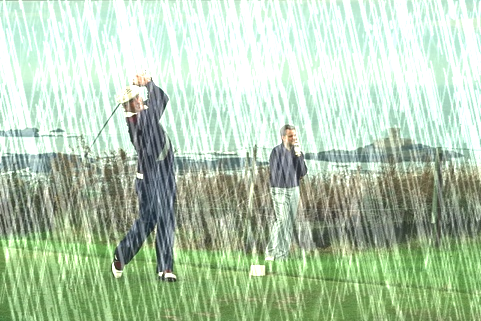}  }
	\subfloat[Background]{\includegraphics[width=.5\columnwidth]{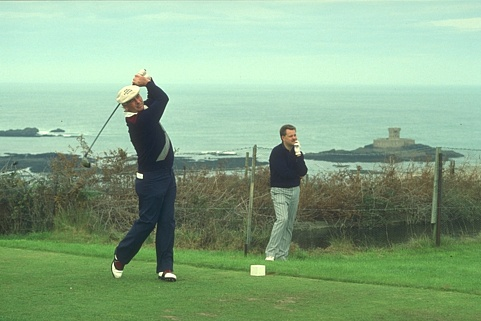}  }
	\subfloat[JORDER\cite{cnn_joint_2017}]{\includegraphics[width=.5\columnwidth]{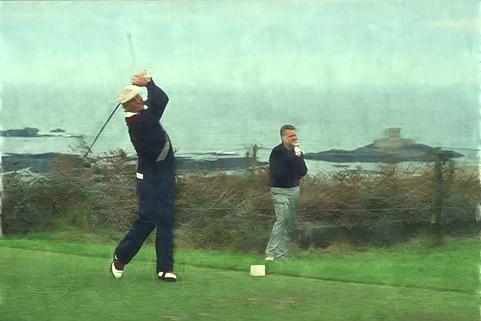} }  \\
	\subfloat[RESCAN\cite{cnn_squeeze_2018}]{\includegraphics[width=.5\columnwidth]{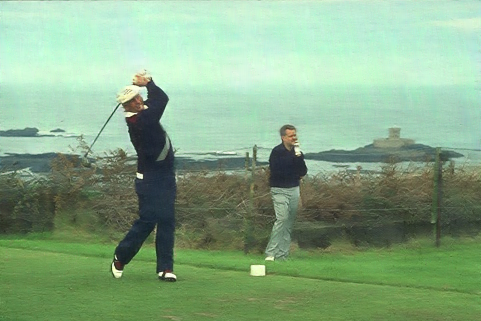} } 
	\subfloat[PReNet\cite{cnn_progressive_2019}]{\includegraphics[width=.5\columnwidth]{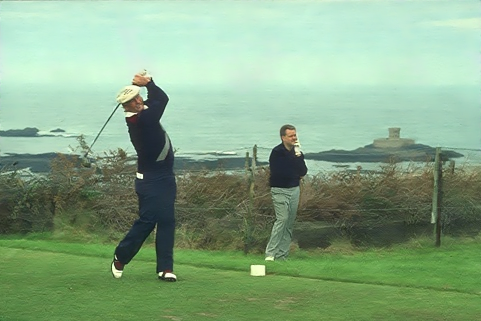} } 
	\subfloat[Our NCANet]{\includegraphics[width=.5\columnwidth]{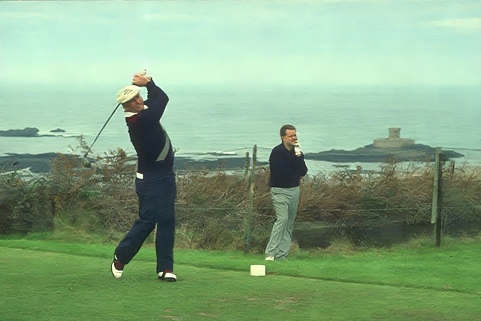} } 
	
	\caption{ 
		Qualitative comparison among NCANet and other state-of-the-art methods on images from Rain100H.
	}
	\label{fig:rain100H}
\end{figure*}

\subsection{Experiments on Rain100L Dataset}

We first evaluate our NCANet on Rain100L Dataset \cite{cnn_joint_2017}, which models the light rain conditions. 
Rain100L contains 200 training images and 100 testing images, and we will train our models as well as other state-of-the-art deraining models for comparison.

\begin{table}[!hb]
	\centering
	\caption{Quantitative comparison among different state-of-the-art deraining models as well as our NCANet on Rain100L dataset \cite{cnn_joint_2017}}
	\vspace{2ex}
	\resizebox{0.5\columnwidth}{!}
	{
		\begin{tabular}{c|c|c}
			\hline \hline
			\tabincell{c}{Model}     &
			\tabincell{c}{PSNR}      &
			\tabincell{c}{SSIM}      \\
			\hline
			GMM \cite{image_layer_2016}          & 28.66       & 0.865         \\
			DDN \cite{cnn_detail_2017}           & 32.16       & 0.936         \\
			RGN \cite{cnn_residual_2018}         & 33.16       & 0.963         \\
			JORDER \cite{cnn_joint_2017}         & 36.61       & 0.974         \\
			PreNet \cite{cnn_progressive_2019}   & 37.48       & 0.979         \\
			NCANet		 				 	     & \textbf{38.506}      & \textbf{0.982}         \\
			\hline \hline
		\end{tabular} 
	}
	\label{table:Rain100L}
	
\end{table}

Fig.~\ref{fig:rain100l} and Fig.~\ref{fig:rain11} show the visual quality of our NCANet compared to other state-of-the-art deraining models on Rain100L dataset. 
It is clear to see from the figures that our model has better performance in terms of measuring qulitatively.
As shown in Table~\ref{table:Rain100L}, our NCANet outperforms all other deraining model by a large margin.
The superiority performance of our model is due to that we insert an NCA block which could capture the long-range spatial and channel dependencies, providing rich information for deraining.

\subsection{Experiments on Rain100H Dataset}

We then evaluate our NCANet on Rain100H Dataset \cite{cnn_joint_2017}, which models the heavy rain conditions. 
Rain100H contains 1800 training samples and 100 testing images; however, we will exclude 546 images from Rain100H dataset and train our models as well as other state-of-the-art deraining models on the rest 1254 images, as these 546 images have the same background with the testing images \cite{cnn_progressive_2019}.

\begin{table}[!ht]
	\centering
	\caption{Quantitative comparison different state-of-the-art deraining models as well as our NCANet on Rain100H dataset \cite{cnn_joint_2017}}
	\vspace{2ex}
	\resizebox{0.5\columnwidth}{!}
	{
		\begin{tabular}{c|c|c}
			\hline \hline
			\tabincell{c}{Model}     &
			\tabincell{c}{PSNR}      &
			\tabincell{c}{SSIM}      \\
			\hline
			GMM \cite{image_layer_2016}          & 15.05         & 0.425         \\
			DDN \cite{cnn_detail_2017}           & 21.92         & 0.764         \\
			RGN \cite{cnn_residual_2018}         & 25.25         & 0.841         \\
			JORDER \cite{cnn_joint_2017}         & 26.54         & 0.835         \\
			RESCAN \cite{cnn_squeeze_2018}	     & 28.88         & 0.866         \\
			PreNet \cite{cnn_progressive_2019}   & 29.46         & 0.899         \\
			NCANet		 				 	     & \textbf{29.923}       & \textbf{0.903}  \\
			\hline \hline
		\end{tabular} 
	}
	\label{table:Rain100H}
	
\end{table}

Fig.~\ref{fig:rain100H} shows the visual quality of our NCANet compared to other state-of-the-art deraining models on Rain100H dataset. 
Rainy images in Rain100H have much heavier rain densities than rainy images in Rain100L, as shown in Fig.~\ref{fig:rain100H}. 
Also, we can see from Table~\ref{table:Rain100H} that our NCANet outperforms other deraining model significantly.
Though as mentioned in the last subsection, by equipped with NCA block, our model could non-locally capture the long-range channel dependencies. And another important factor for our NCANet is because we have an iterative rain model in Eqn.~\eqref{eqn:new_blend2} and adopt recurrent training scheme as shown in Fig.~\ref{fig:framework}. That makes to be powerful in have rain.

\subsection{Experiments on Rain1400 Dataset}

We conduct further experiment on another dataset \cite{cnn_detail_2017}, which contains 12600 samples for training and 1400 samples for testing (a.k.a Rain1400).
This dataset is more complicated than Rain100L and Rain100H, there are 14 different rain directions in Rain1400 dataset.
Different rain directions and heavy rain streaks impose difficulty in both model training and testing.
However, due to effectiveness of our rain model in Eqn.~\eqref{eqn:new_blend2} and our novel NCA block, our NCANet have a outperform improvement in  difficult scenarios (more details in Table~\ref{table:Rain1400}).

\begin{table}[!ht]
	\centering
	\caption{Quantitative comparison different state-of-the-art deraining models as well as our NCANet on Rain1400 dataset \cite{cnn_detail_2017}}
	\vspace{2ex}
	\resizebox{0.5\columnwidth}{!}
	{
		\begin{tabular}{c|c|c}
			\hline \hline
			\tabincell{c}{Model}     &
			\tabincell{c}{PSNR}      &
			\tabincell{c}{SSIM}      \\
			\hline
			GMM \cite{image_layer_2016}          & 23.21       & 0.803         \\
			DDN \cite{cnn_detail_2017}           & 29.91         & 0.910      \\
			RGN \cite{cnn_residual_2018}         & 30.76       & 0.929         \\
			JORDER \cite{cnn_joint_2017}         & 31.58       & 0.937         \\
			PreNet \cite{cnn_progressive_2019}   & 32.60         & 0.946         \\
			NCANet		 				 	     & \textbf{32.78}       & \textbf{0.947}  \\
			\hline \hline
		\end{tabular} 
	}
	\label{table:Rain1400}
	
\end{table}

\subsection{Computation and memory footprint of NCA Block}

 NCL block is memory-intensive to implement the tensor multiplication $HWC\times CHW$ for the attention map $HW\times HW$. For NCA block, although it generates more attention maps: $C\times C$, $H\times H$ and   $W\times W$, the memory size and dimension of computation are reduced. As shown in Table~\ref{table:Memory}, The NCA Block save more memory than NCL Block.
 
 \begin{table}[!ht]
 	\centering
 	\caption{The memory comparison between the NCL Block and NCA Block. And the results in the table are the additional values after adding various modules.}
 	\vspace{2.3ex}
 	\resizebox{0.7\columnwidth}{!}
 	{
 		\begin{tabular}{c|c|c|c}
 			\hline \hline
 			\tabincell{c}{Model}     &
 			\tabincell{c}{Memory(M)}      &
 			\tabincell{c}{FLOPs}      &
 			\tabincell{c}{Time}      \\
 			\hline
 			baseline               & 4981 & 10.16GMac  & 0.5h  \\
 			+NCL \cite{nonlocal2018} & 16274 & 0.74GMac & 1.0h \\
			+scSE \cite{Roy2018Concurrent} & 2197 & 0.62GMac & 0.4h \\	
 			+NCA                    & 1740 & 0.57GMac   & 0.2h  \\
 			\hline \hline
 		\end{tabular} 
 	}
 	\label{table:Memory}
 	
 \end{table}

\subsection{Recursive Number of stage T}

Table~\ref{table:4} lists the PSNR and SSIM values which are derived by using our NCANet models with stages $T$ = 3,4,5,6,7. From this it can be seen that NCANet with more recursive stages (from 3 to 6) usually leads to higher values. However, having a larger $T$ also makes NCANet more difficult to de-convert. When $T$ = 7, $\text{NCANet}_7$ performs a little inferior to $\text{NCANet}_6$. Thus, in the following experiments we set $T$ = 6 to replace.

\begin{table}[!ht]
	\caption{Comparison of NCANet models with different $T$ stages.}
	\centering
	\begin{tabular}{c|c|c|c|c|c}
		\hline \hline
		\tabincell{c}{$\text{NCANet}_T$}     &
		\tabincell{c}{$3$}     &
		\tabincell{c}{$4$}      &
		\tabincell{c}{$5$}      &
		\tabincell{c}{$6$}      &
		\tabincell{c}{$7$}      \\
		\hline
		PSNR	& $37.103$ & $37.982$ & $38.264$	& $38.506$ & $38.075$  \\
		SSIM	& $0.973$ & $0.976$ & $0.980$	& $0.982$ & $0.979$ \\
		\hline \hline
	\end{tabular}
	\label{table:4}
	\vspace{-0.2cm}
\end{table}

\subsection{The order of the three sub-blocks}

We have changed the order of three sub-blocks in the residual block for different position, psnr/ssim seems to be no significant difference as shown in the additional Table~\ref{table:order}.

\begin{table}[t!]
	\caption{The order of three sub-blocks.}
	\centering
	\begin{tabular}{c|c|c|c|c|c|c  }
		\hline \hline
		VA	& $1$ & $1$ & $2$	& $3$ & $2$ & $3$ \\
		\hline
		TA	& $2$ & $3$ & $1$	& $1$ & $3$ & $2$ \\
		\hline
		CA	& $3$ & $2$ & $3$ & $2$ & $1$ & $1$  \\
		\hline
		PSNR   & $38.506$ & $38.498$ & $38.457$	& $38.464$ & $38.417$ & $38.376$ \\
		SSIM	& $0.982$ & $0.981$ & $0.981$	& $0.981$ & $0.981$  & $0.980$ \\
		\hline \hline
	\end{tabular}
	\label{table:order}
	\vspace{-0.2cm}
\end{table}

\section{conclusions}

\label{sec:conclusions}

In this paper, we propose a novel non-local channel aggregation network, the main contribution of the proposed method lies in that to the deraining field. 
First, unlike previous plain rain layer separation rain model, we build a new rain model, which could capture more detailed information of rainy images.
Second, based on the new rain model, we develope a novel recurrently connected residual nets.
The most important thing is that we develope a new attention block called non-local channel aggregation block, which could capture the long-range spatial and channel dependencies simultaneously.
At last, experimental results on public datasets shows that,
the proposed method can be applied in the low-level computer vision field.

{
	\bibliographystyle{ieee}
	\bibliography{egbib}
}

\vspace{-40 mm}
\begin{IEEEbiography}[{\includegraphics[width=1in,height=1.25in,clip]{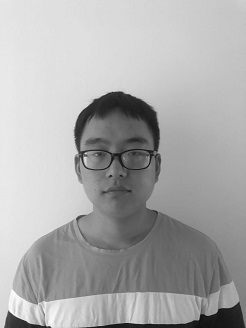}}]{Zhipeng Su}
received the B.S. degree in electronic information engineering from Fuzhou University, Fuzhou, China, in 2016, He is currently pursuing the Ph.D. degree in signal and information processing in xiamen University, Xiamen, China. He current research interests include pattern recognition, image processing, and computer vision.
\end{IEEEbiography}
\vspace{-40 mm}
\begin{IEEEbiography}[{\includegraphics[width=1in,height=1.25in,clip]{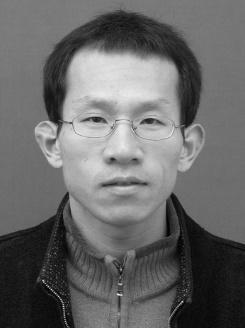}}]{Yixiong Zhang}
was born in Fujian, China, in 1981. He received the B.S. degree in information engineering and the Ph.D. degree in information and communication engineering from Zhejiang University, Hangzhou, China, in 2003 and 2009, respectively. In 2009, he joined the School of Informatics, Xiamen University, Xiamen, China, where he is currently an Associate Professor. His current research interests include image/video processing, signal detection and parameter estimation, radar imaging, and hardware/software co-design of embedded systems.
\end{IEEEbiography}
\enlargethispage{-3cm}
\begin{IEEEbiography}[{\includegraphics[width=1in,height=1.25in,clip]{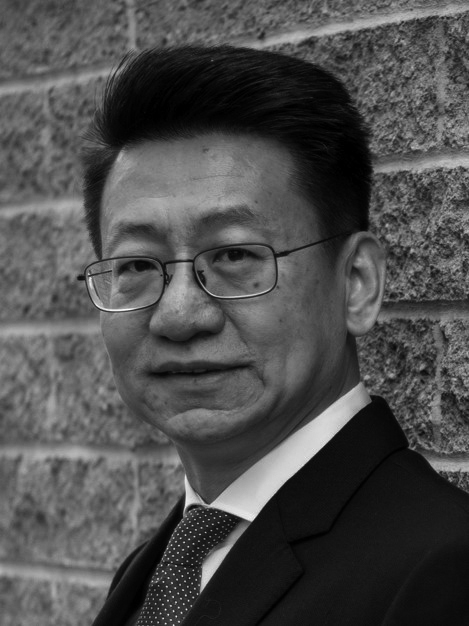}}]{Xiao-Ping Zhang}
received B.S. and Ph.D. degrees from Tsinghua University, in 1992 and 1996, respectively, both in Electronic Engineering. He holds an MBA in Finance, Economics and Entrepreneurship with Honors from the University of Chicago Booth School of Business, Chicago, IL. 

Since Fall 2000, he has been with the Department of Electrical, Computer and Biomedical Engineering, Ryerson University, Toronto, ON, Canada, where he is currently a Professor and the Director of the Communication and Signal Processing Applications Laboratory. He has served as the Program Director of Graduate Studies. He is cross-appointed to the Finance Department at the Ted Rogers School of Management, Ryerson University. He was a Visiting Scientist with the Research Laboratory of Electronics, Massachusetts Institute of Technology, Cambridge, MA, USA, in 2015 and 2017. He is a frequent consultant for biotech companies and investment firms. He is the Co-Founder and CEO for EidoSearch, an Ontario-based company offering a content-based search and analysis engine for financial big data. His research interests include statistical signal processing, machine learning, image and multimedia content analysis, sensor networks and IoT, and applications in big data, finance, and marketing. 

Dr. Zhang is Fellow of the Canadian Academy of Engineering, Fellow of the Engineering Institute of Canada, Fellow of the IEEE, a registered Professional Engineer in Ontario, Canada, and a member of Beta Gamma Sigma Honor Society. He is the general Co-Chair for the IEEE International Conference on Acoustics, Speech, and Signal Processing, 2021. He is the general co-chair for 2017 GlobalSIP Symposium on Signal and Information Processing for Finance and Business, and the general co-chair for 2019 GlobalSIP Symposium on Signal, Information Processing and AI for Finance and Business. He is an elected Member of the ICME steering committee. He is the General Chair for the IEEE International Workshop on Multimedia Signal Processing, 2015. He is the Publicity Chair for the International Conference on Multimedia and Expo 2006, and the Program Chair for International Conference on Intelligent Computing in 2005 and 2010. He served as a Guest Editor for Multimedia Tools and Applications and the International Journal of Semantic Computing. He was a tutorial speaker at the 2011 ACM International Conference on Multimedia, the 2013 IEEE International Symposium on Circuits and Systems, the 2013 IEEE International Conference on Image Processing, the 2014 IEEE International Conference on Acoustics, Speech, and Signal Processing, the 2017 International Joint Conference on Neural Networks and the 2019 IEEE International Symposium on Circuits and Systems. He is Senior Area Editor for the IEEE TRANSACTIONS ON SIGNAL PROCESSING and the IEEE TRANSACTIONS ON IMAGE PROCESSING. He was Associate Editor for the IEEE TRANSACTIONS ON IMAGE PROCESSING, the IEEE TRANSACTIONS ON MULTIMEDIA, the IEEE TRANSACTIONS ON CIRCUITS AND SYSTEMS FOR VIDEO TECHNOLOGY, the IEEE TRANSACTIONS ON SIGNAL PROCESSING, and the IEEE SIGNAL PROCESSING LETTERS. He received 2020 Sarwan Sahota Ryerson Distinguished Scholar Award the Ryerson University highest honor for scholarly, research and creative achievements. He is selected as IEEE Distinguished Lecturer by the IEEE Signal Processing Society for the term 2020 to 2021, and by the IEEE Circuits and Systems Society for the term 2021 to 2022.  Business, Chicago, IL.
\end{IEEEbiography}

\begin{IEEEbiography}[{\includegraphics[width=1in,height=1.25in,clip]{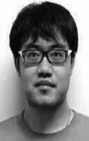}}]{Feng Qi}
did bachelor studies in Zhejiang University and got his master and PhD degrees from Katholieke Universiteit Leuven, Belgium in 2005 and 2011, respectively. Since 2005, he was working on antennas and millimeter wave imaging. Then he joined RIKEN, Japan, by working on nonlinear optics. Since 2014, he was doing research on THz radar in Goethe University, Germany and University of Birmingham, UK. In 2015, he joined Chinese Academy of Sciences as a professor by heading the Terahertz Imaging Lab in Shenyang Institute of Automation. Since 2012, he served the Global Symposium on Millimeter Waves as a TPC member and he got the “Best Paper Award” in 2014. Now his research interests include: microwave laser and radar.
\end{IEEEbiography}
\end{document}